\pdfoutput=1

\documentclass[11pt]{article}

\usepackage[preprint]{acl}

\usepackage{times}
\usepackage{latexsym}

\def\ModelName{COgnitive REasoning in Movies}
\def\DatasetName{MovieCORE}

\def\MethodName{ACE}
\def\MethodNameExt{Agentic Choice Enhancement}

\usepackage{multirow}
\usepackage{wrapfig}
\usepackage{amsmath} 
\usepackage{adjustbox}
\usepackage{algorithm}
\usepackage{algpseudocode}
\usepackage{cleveref}
\usepackage{booktabs}
\usepackage{makecell} 
\usepackage{morewrites}
\usepackage{pdfpages}
\usepackage{url}
\usepackage{hyperref} 

\usepackage[T1]{fontenc}

\usepackage[utf8]{inputenc}

\usepackage{microtype}

\usepackage{inconsolata}

\usepackage{graphicx}

\title{\DatasetName: \ModelName}

\author{
\textbf{Gueter Josmy Faure}$^{1}$,  
\textbf{Min-Hung Chen}$^{2}$, 
\textbf{Jia-Fong Yeh}$^{1}$,
\textbf{Ying Cheng}$^{3}$,
\textbf{Hung-Ting Su}$^{1}$, \\
\textbf{Yung-Hao Tang},
\textbf{Shang-Hong Lai}$^{3}$,
\textbf{Winston H. Hsu}$^{1}$ \\ \\
$^{1}$National Taiwan University, $^{2}$NVIDIA, $^{3}$National Tsing Hua University\\
}

\begin{document}

\twocolumn[{%
	\renewcommand\twocolumn[1][]{#1}%
	\maketitle
	\begin{center}
		\newcommand{\teaserwidth}{\textwidth}
		\centerline{
			\includegraphics[width=\teaserwidth,clip]{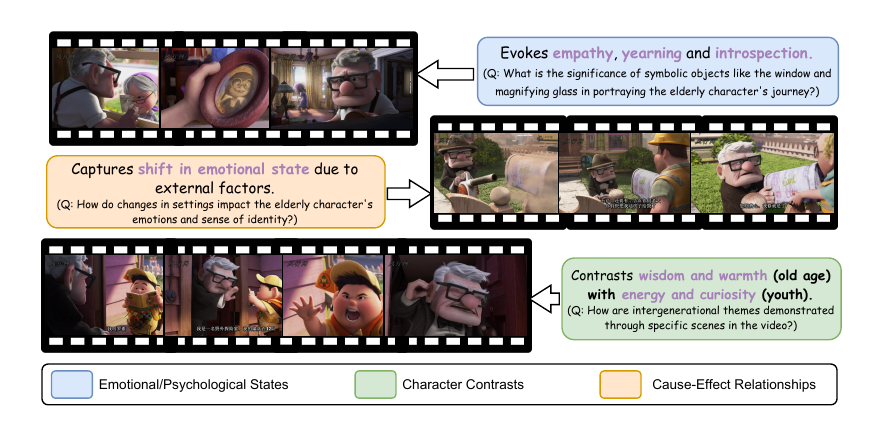}
		}
		\captionof{figure}{\textbf{Beyond Shallow Video Understanding:} The proposed benchmark, \DatasetName{}, challenges vision-language models (VLMs) to understand the subtle interplay between emotions \textit{(Top, Middle)}, character dynamics and causality \textit{(Middle, Bottom)}, and psychological complexity \textit{(Top, Middle)}. From empathy to introspection, from wisdom to curiosity \DatasetName{} tests VLMs' ability to comprehend the deeper elements of movies.}
		\vspace{-0.05in}
		\label{fig:teaser}
	\end{center}%
}]

\begin{abstract}
    
    This paper introduces \DatasetName, a novel video question answering (VQA) dataset designed to probe deeper cognitive understanding of movie content. Unlike existing datasets that focus on surface-level comprehension, \DatasetName\ emphasizes questions that engage System-2 thinking while remaining specific to the video material. We present an innovative agentic brainstorming approach, utilizing multiple large language models (LLMs) as thought agents to generate and refine high-quality question-answer pairs. To evaluate dataset quality, we develop a set of cognitive tests assessing depth, thought-provocation potential, and syntactic complexity. We also propose a comprehensive evaluation scheme for assessing VQA model performance on deeper cognitive tasks. To address the limitations of existing video-language models (VLMs), we introduce an agentic enhancement module, \MethodNameExt\ (\MethodName), which improves model reasoning capabilities post-training by 25\%. Our work contributes to advancing movie understanding in AI systems and provides valuable insights into the capabilities and limitations of current VQA models when faced with more challenging, nuanced questions about cinematic content. Our project page, dataset and code can be found at \url{https://joslefaure.github.io/assets/html/moviecore.html}
\end{abstract}

\section{Introduction}
\label{sec:intro}
Movie audiences consciously or subconsciously absorb information about actors' states of mind, body language, and expressions to infer their moods and empathize with their situations. Most people would agree that such inferences are crucial to truly understanding a movie. Despite the significance of this deeper level of understanding, existing movie-based VQA datasets have yet to explore this aspect of film comprehension.

Recent movie-based VQA datasets \cite{wu2021towards, song2024moviechat, rawal2024cinepile} primarily focus on surface-level understanding, neglecting the challenge of comprehending movies at a deeper cognitive level. They predominantly address the ``what'' by posing questions such as ``What is the relationship between the actors?'' or ``What time does the video take place?'', and largely overlook the ``how,'' ``why,'' and ``why not'' questions crucial for achieving a profound understanding of movies. While EgoSchema \cite{mangalam2023egoschema} attempts to delve beyond the obvious, its more profound questions often remain general.

We propose \DatasetName, a novel VQA dataset designed to engage System-2 thinking—the slow, deliberate, and logical cognitive processes—while maintaining strict relevance to specific video content. Unlike existing datasets, \DatasetName~embraces the inherent subjectivity of "why" and "why not" questions as a feature rather than a limitation, creating both meaningful challenges and research opportunities. To generate comprehensive and faithful question-answer pairs, we develop an agentic brainstorming approach that leverages multiple large language models (LLMs) as interactive thought agents that engage in continuous discussions to refine QA pairs. We validate the quality of the QAs through rigorous human review of a representative subset. Additionally, we employ quantitative cognitive metrics to measure our dataset's depth and syntactic complexity relative to existing benchmarks. Our evaluation of current VQA models on \DatasetName reveals critical insights about their performance on these challenging cognitive tasks. To address identified limitations and improve existing VLMs' deeper cognitive reasoning capabilities, we introduce \MethodNameExt~(\MethodName), which demonstrates relative performance improvements of up to 25\% compared to baseline approaches.

Our key contributions are the following:
\begin{itemize}
    \item We introduce \DatasetName, a VQA dataset focused on thought-provoking questions and answers specific to movie content.
    \item We develop an agentic brainstorming approach using multiple LLMs as agents to generate and refine high-quality QA pairs.
    \item We implement a set of cognitive tests to evaluate the depth, thought-provocation, and complexity of VQA datasets.
    \item We design a comprehensive evaluation scheme to assess the accuracy, comprehensiveness, depth, and coherence of answers from existing Vision Language Models (VLMs).
    \item We evaluate several VLMs on our dataset in both zero-shot and fully-supervised settings, offering insights into their performance on deeper cognitive tasks.
    \item We propose a post-training "agentic selection" plugin to improve existing VLMs and show a relative improvement of up to 25\% compared to the baseline.
\end{itemize}

\begin{figure*}[ht!]
    \centering
    \includegraphics[width=\textwidth]{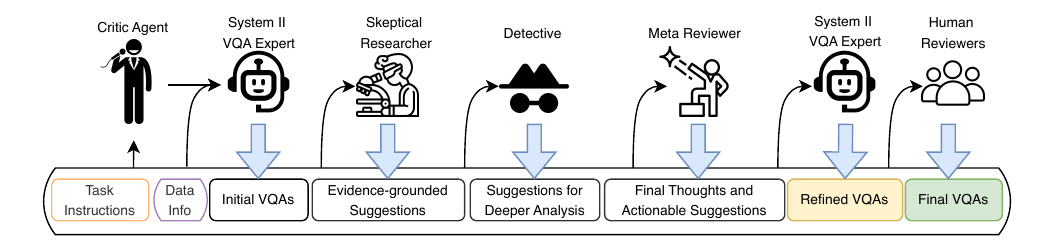}
    \caption{The \textit{Critic Agent}, acting as the master of ceremonies (MC), orchestrates interactions among specialized agents using video context and task instructions. It sequentially engages the \textit{System II VQA Expert}, \textit{Skeptical Researcher}, \textit{Detective}, and \textit{Meta Reviewer}, accumulating insights at each stage. Upon receiving final recommendations from the \textit{Meta Reviewer}, the MC relays them to the \textit{System II VQA Expert} for VQA refinement. Subsequently, a subset of these refined VQAs undergoes evaluation by human experts for final validation.}
    \label{fig:workflow}
\end{figure*}

\section{Related Work}
\label{sec:related_work}

\noindent \textbf{Movie-Based Question-Answering Datasets.}
Recent video understanding benchmarks are often based on movie scenes because films offer a rich blend of multimodal content, combining visual, linguistic, and temporal elements within complex narratives. Early efforts like MovieQA \cite{tapaswi2016movieqa} explores entire movie understanding but were limited by questions heavily relying on dialogue. TVQA \cite{lei2018tvqa} requires reasoning over multiple events in short TV series clips, integrating visuals and subtitles. LVU \cite{wu2021towards} addresses scaling video comprehension to extended sequences, necessitating models to process long temporal contexts. MAD \cite{soldan2022mad} and its extension \cite{han2023autoad} focus on scene-level descriptions through audio and visuals but were mainly used for scene annotation tasks with limited narrative comprehension. MoVQA \cite{zhang2023movqa} introduces multi-level questions, challenging models in temporal perception, causal reasoning, and narrative synthesis. CinePile \cite{rawal2024cinepile} automates large-scale question generation across varied scenes and question type and MovieChat-1k \cite{song2024moviechat} focuses on basic understanding of cinematic contexts. 

\noindent \textbf{Video Question-Answering Reasoning.} Text-based reasoning datasets like DROP \cite{dua2019drop} and GSM8K \cite{cobbe2021training} handle discrete reasoning tasks, including counting and arithmetic, but are limited to textual inputs and do not address the complexities involved in integrating visual reasoning. Egocentric datasets, such as EpicKitchens \cite{damen2018scaling}, Ego4D \cite{grauman2022ego4d}, and EgoSchema \cite{mangalam2023egoschema}, challenge models to interpret subjective interactions and continuous activities from a first-person perspective, requiring both perceptual understanding and intention reasoning. Perception Test \cite{patraucean2024perception} broadens perceptual reasoning to varied video contexts, assessing high-level reasoning abilities. Multi-task and complex video benchmarks, such as MVBench \cite{li2024mvbench}, Video-MME \cite{fu2024video}, and MLVU \cite{zhou2024mlvu}, integrate multiple reasoning challenges, requiring predictive reasoning, memory recall, and cross-modal inference over long video sequences. While these datasets have advanced various aspects of video understanding, they predominantly rely on surface-level comprehension of video content. Our work introduces the first dataset specifically designed to evaluate System-2 reasoning in the video domain, requiring models to engage in slow, deliberate, and analytical thinking processes aiming to mirror human approaches to complex movie understanding.

\section{\DatasetName~ Creation and Curation}
To address the challenge of obtaining question-answer pairs that delve into deeper levels of movie understanding, we propose an agentic annotation workflow. This approach leverages the deliberative capabilities of multiple LLMs acting as specialized agents, each contributing unique perspectives to the annotation process. We start with video context extraction to make sure our text-only annotation agents have enough information about the video.

\subsection{Video Context Extraction}
\label{subsect:context}
The videos for our dataset are sourced from MovieChat-1k \cite{song2024moviechat}, a collection of 1,000 movie clips averaging 10 minutes each. We use 986 of these clips, as 14 were either unavailable or lacked necessary annotations. MovieChat-1k, already provides high-level information for each video, such as temporal setting (e.g., ancient or modern) and metadata like the movie's genre. Although some videos in the original dataset include captions, we observe inaccuracies and imbalanced descriptions. Therefore, we exclude these captions, focusing instead on the existing QA pairs and movie metadata.

To provide video context, we utilize MiniCPM-v2.6 \cite{yao2024minicpm}, an open-source model with visual capabilities comparable to GPT-4V. We prompt it with a carefully curated set of eight questions (shown in \Cref{fig:prior} in the supplementary material) designed to extract a multi-dimensional understanding of the video. These questions address narrative structure, thematic focus, emotional tone, key events, character dynamics, genre, and target audience. The extracted information serves as \textit{Data Info} priors for our agents.

\begin{figure*}[!t]
    \centering
    \includegraphics[width=\textwidth]{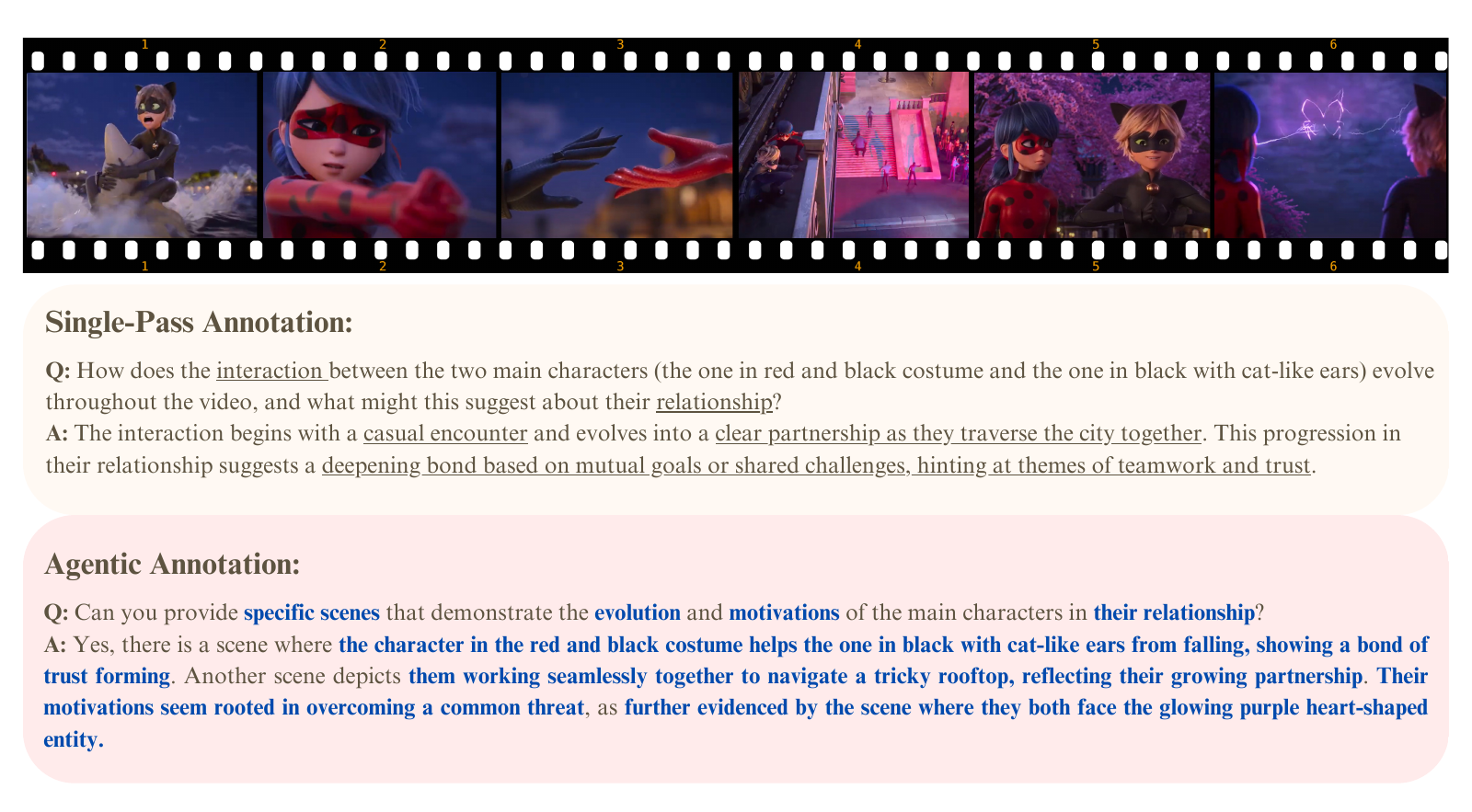}
    \caption{\textbf{Comparison of single-pass and agentic annotation}. The agentic method (bottom) elicits specific scene details, concrete examples, and detailed story elements, demonstrating the enhanced granularity achieved through multi-agent refinement. Text in blue indicates new, specific details absent in the single-pass version. The single-pass annotation (top), on the other hand, while also attempting to ask deeper questions, remains at a more abstract level.}
    \label{fig:agentic_comparison1}
\end{figure*}

\subsection{Agentic Annotation Workflow}
\label{subset:workflow}

Our workflow, illustrated in \Cref{fig:workflow}, employs a multi-agent system orchestrated by a Critic Agent acting as the master of ceremonies (MC). Using the Agentic AI framework autogen \cite{wu2023autogen}, we deploy instances of GPT4-o for the VQA Expert and Meta Reviewer roles (as these positions demand superior reasoning capabilities), with GPT4-o-mini powering the other expert agents. The process begins as the Critic Agent receives task instructions and video context (\textit{Data Info}) extracted as described in \Cref{subsect:context} and sends them to the System II VQA Expert who generates questions that engage System-2 thinking. These initial QA pairs are then scrutinized by the Skeptical Researcher, who evaluates their contextual relevance and accuracy, often challenging the VQA Expert to provide more concrete evidence. The Detective agent follows, suggesting additional questions to uncover underlying motivations and biases. The Meta Reviewer synthesizes these insights, proposing enhancements to the initial VQAs. The Critic Agent then consolidates this feedback for the VQA Expert to refine the QAs. The process concludes with human expert evaluation of a subset of the refined VQAs, assessing their clarity, depth, relevance, and answerability. This agentic annotation workflow mimics collaborative human expert discussions by harnessing collective intelligence and mitigating potential biases of any single agent\footnote{Wondering why we chose these specific agents? Please see \Cref{subsect:agentic_singlepass_comp} and \ref{subsect:why}}. 

To ensure the quality and reliability of our dataset, we implement a rigorous human verification process. Seven graduate students were recruited to assess a subset of 30 videos, 30 captions and 150 QA pairs. The final human validation ensures that the resulting VQAs meet the highest standards of quality and depth. We provide more details on the human validation in \Cref{sec:human}.

\begin{table*}[!t]
\centering
    \centering
    \small
    \begin{tabular}{lcccccccc}
        \toprule
        \multirow{2}{*}{\textbf{Dataset}} & \multicolumn{3}{c}{\textbf{Parse Tree Depth}} & \multicolumn{3}{c}{\textbf{F–K Grade Score}} & \multirow{2}{*}{\textbf{BT Level}} & \multirow{2}{*}{\textbf{HO-QA (\%)}} \\
        \cline{2-7}
         & Q & A & Avg & Q & A & Avg & & \\
        \midrule
        MovieChat-1k \cite{song2024moviechat} & 3.58 & 1.31 & 2.45 & 3.19 & -0.39 & 1.4 & 1.8 & 0.0 \\
        ActivityNetQA \cite{yu2019activitynet} & 4.24 & 0.27 & 2.26 & 2.69 & 0.98 & 1.84 & 1.9 & 0.2 \\
        MVBench \cite{li2024mvbench} & 3.96 & 1.71 & 2.84 & 4.74 & 1.47 & 3.11 & 2.2 & 3.4 \\
        EgoSchema \cite{mangalam2023egoschema}  & \textbf{6.56} & \underline{4.38} & \underline{5.47} & \underline{10.52} & \underline{6.08} & \underline{8.30} & \underline{3.1} & \underline{33.1} \\
        \midrule
        \textbf{\DatasetName} & \underline{5.38} & \textbf{6.39} & \textbf{5.88} & \textbf{12.98} & \textbf{15.07} & \textbf{14.03} & \textbf{4.9} & \textbf{99.2} \\
        \bottomrule
    \end{tabular}
    \caption{\textbf{Syntactic Complexity and Cognitive Demand Comparison:} Parse tree depth, Flesch-Kincaid (F-K) grade scores, average Bloom's Taxonomy (BT) level, and percentage of higher-order questions and answers (HO-QA) across various VQA datasets. Q and A represent questions and answers respectively. Best results are in \textbf{bold}, second-best are \underline{underlined}.}
    \label{tab:syntactic}
\end{table*}

\subsection{Agentic versus Single-Pass Annotation}
To illustrate the effectiveness of the proposed Agentic Annotation workflow, we compare the quality of the VQAs generated by the System II VQA Expert in the initial round (single-pass) and those produced through our workflow after the agent has gathered feedback and enhancement ideas from other experts (agentic annotation). As shown in \Cref{fig:agentic_comparison1}, the agentic annotation approach demonstrates clear advantages over single-pass annotation. While the single-pass annotation provides a general, abstract description of character relationships, the agentic annotation generates questions that ask for and answers that deliver specific, concrete details about key scenes that support the relationship development of the characters -- including the falling scene, rooftop navigation, and confrontation with the purple heart-shaped entity. The agentic process elicits richer context and more granular evidence, making the annotations more specific and faithful to the movie content. It also makes the dataset much more valuable for training and evaluating AI systems' understanding of narrative progression and character dynamics. This suggests that using multiple AI agents as thought partners leads to more detailed and substantive annotations compared to traditional single-pass methods used by other auto-annotated datasets such as \cite{rawal2024cinepile} and \cite{mangalam2023egoschema}. More comparisons between agentic and single-pass annotation can be found in \Cref{subsect:agentic_singlepass_comp}.

\subsection{Dataset Description}
\begin{figure}[!ht]
    \centering
    \includegraphics[width=0.5\textwidth]{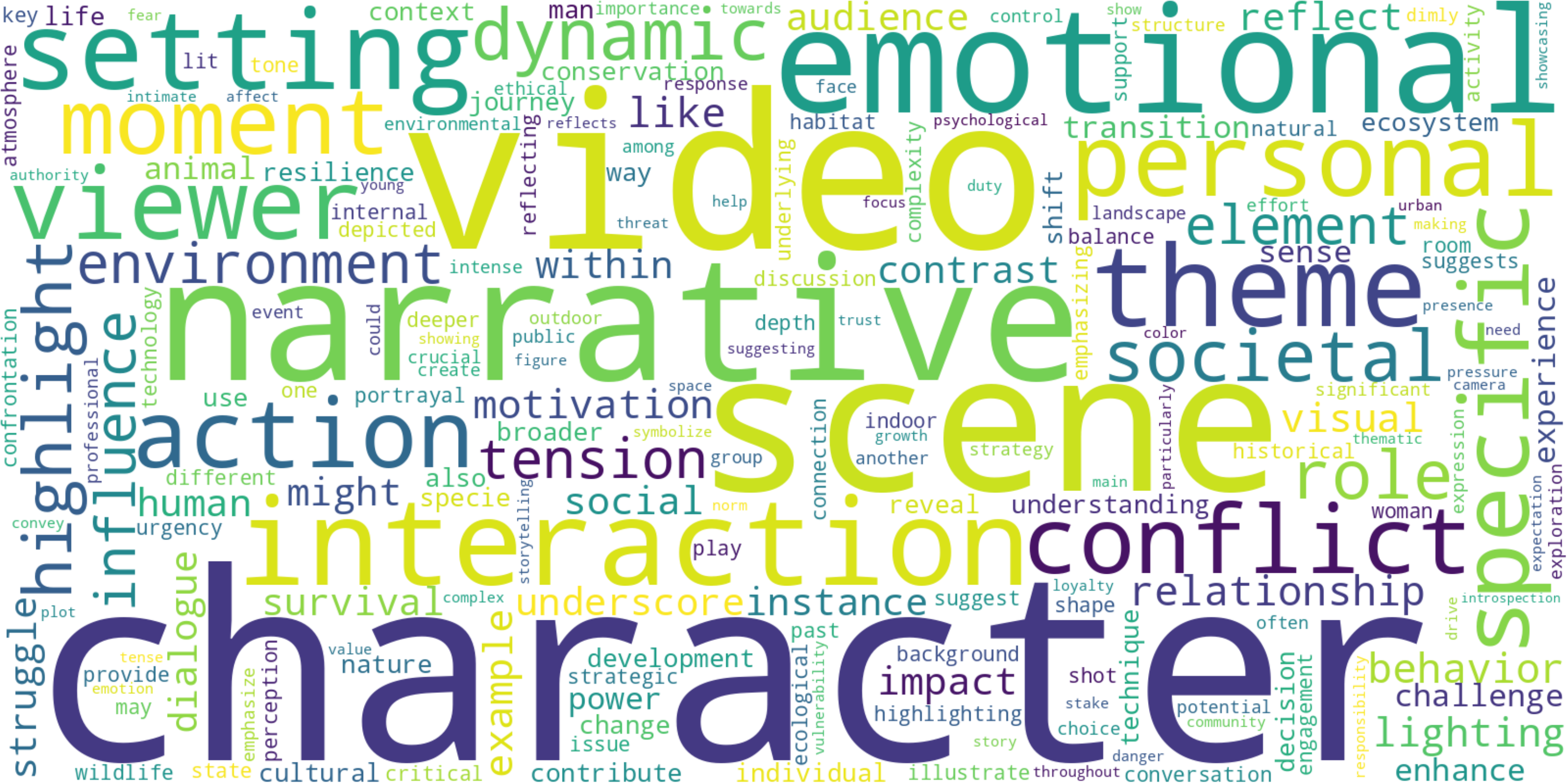}
    \caption{Wordcloud illustrating key themes and concepts of \DatasetName~with terms such as "emotional", "character" and "influence" very prominent.}
    \label{fig:wordcloud}
\end{figure}

\DatasetName\ is a video question-answering (VQA) dataset designed to probe deeper cognitive understanding of movie content. The dataset comprises 986 videos paired with 4,930 corresponding questions and answers and 986 captions. Following the splits of the original MovieChat-1k dataset \cite{song2024moviechat}, we split MovieCORE into 4080 QAs for training (816 videos) and 850 for testing (170 videos).
The primary application of \DatasetName\ lies in training and evaluating VQA models' capabilities in deeper cognitive tasks. The questions are specifically designed to assess models' abilities to comprehend complex narrative elements, character motivations, and subtle contextual cues -- skills that are crucial for achieving human-like understanding of cinematic content. A wordcloud of \DatasetName~ is shown in \Cref{fig:wordcloud} suggesting complex themes regarding character dynamics, emotional resonance, and societal implications through terms like ``tension,'' ``psychological,'' ``cultural,'' and ``emotional.'' Also, the prominence of analytical terms such as ``underscore'',``depth,'' and ``critical,'' suggests questions that probe deeper interpretations and thematic elements rather than just literal plot descriptions. 

\section{Experiments}
\label{sec:exp}
\subsection{Linguistic and Cognitive Complexity}
To evaluate the effectiveness of \DatasetName{} in engaging System-2 thinking and promoting deeper cognitive processing, we conduct a series of tests designed to assess the complexity, readability, and cognitive demand of our questions and answers. These tests include well-established metrics such as parse tree depth, Flesch-Kincaid grade score, and Bloom's taxonomy classification. Each provides unique insights into different aspects of our dataset's ability to stimulate higher-order thinking. \Cref{tab:syntactic} presents a comparative analysis of \DatasetName~ against other VQA datasets.

\noindent \textbf{Parse Tree Depth} measures the syntactic complexity of sentences by analyzing their hierarchical structure. We utilize this metric to assess the structural intricacy of our questions and answers. We employ the spaCy library to generate parse trees for each question and answer in our dataset and recursively compute their depth as follows.
Let \( d(t) \) be the depth of a token \( t \) in the tree. For a token with children \( C(t) \), the depth is defined as:
\begin{equation}
    d(t) = 
    \begin{cases} 
    0 & \text{if } C(t) = \emptyset \\
    1 + \max_{c \in C(t)} d(c) & \text{if } C(t) \neq \emptyset 
    \end{cases}
\end{equation}
where \( d(t) = 0 \) if \( t \) is a leaf node (no children), \( d(t) = 1 + \max_{c \in C(t)} d(c) \) if \( t \) has children \( C(t) \), with \( \max_{c \in C(t)} d(c) \) representing the maximum depth of the children of \( t \).
For a sentence with multiple tokens, the depth of the parse tree \( D \) rooted at the token \( r \) (root of the sentence) is \( D = d(r) \).

The depth of these trees are then averaged across the dataset.
A greater parse tree depth often correlates with more complex sentence structures, which typically require more cognitive resources to process. By measuring this, we aim to quantify the linguistic sophistication of our VQAs as compared to existing datasets', hypothesizing that questions and answers with higher parse tree depths are more likely to engage System-2 thinking. \Cref{tab:syntactic} shows that \DatasetName\ has the highest average parse tree depth compared to the other VQA datasets.

\noindent \textbf{The Flesch-Kincaid (F-K) Grade Score} is a readability measure that indicates the U.S. grade level needed to understand a text. We calculate this score for both questions and answers in our dataset using the standard Flesch-Kincaid formula below
\begin{equation}
    \adjustbox{scale=0.9}{$\text{F-K Grade Score} = 0.39 \left( \frac{W}{S} \right) + 11.8 \left( \frac{Y}{W} \right) - 15.59$}
\end{equation}
where \( W \) is the total number of words in the text, \( S \), total number of sentences and \( Y \) the total number of syllables.

While our goal is not to make the content unnecessarily difficult, a moderately high Flesch-Kincaid score indicates that the QAs require a more advanced level of comprehension and thinking. As shown in \Cref{tab:syntactic}, \DatasetName\ substantially outperforms other datasets with an average grade score of 14.03, with its closest competitor -- EgoSchema \cite{mangalam2023egoschema} -- standing at 8.3.

\noindent \textbf{Bloom's Taxonomy} is a hierarchical model used to classify educational learning objectives into levels of complexity and specificity \cite{Mcdaniel_1970}. We prompt GPT-4o-mini with a comprehensive breakdown of the Bloom's Taxonomy and ask it to classify each question and answer into one of six cognitive levels: Remember (1), Understand (2), Apply (3), Analyze (4), Evaluate (5), and Create (6). Such classification helps us assess the cognitive demand of the QAs. Questions falling into higher levels of Bloom's Taxonomy (Analyze, Evaluate, Create) require deeper analysis and critical thinking skills susceptible to trigger System-2 thinking. \DatasetName\ achieves the highest average Bloom Taxonomy Level (BT Level) of 4.9, indicating that our questions and answers predominantly engage higher-order cognitive skills, significantly surpassing the other datasets. Additionally, we report the percentage of higher-order questions and answers (HO-QA), representing the proportion of both questions and answers that fall into the upper levels of Bloom's Taxonomy (levels 4-6). \DatasetName\ excels in this metric with 99.2\% of its questions and answers classified as higher-order.

\begin{algorithm}
\caption{\MethodName{}: \MethodNameExt{}}
\label{alg:hermes_plus_plus}
\begin{algorithmic}[1]
\State \textbf{Input}: Video $V$, Question $Q$, Beam width $k=5$
\State \textbf{Output}: Best response $R^*$
\State $C \gets \text{VLM.generate}(V, Q, \text{beam\_width}=k)$ 
\State $S \gets \text{Llama-3.2.score}(C)$ \Comment{Score candidates}
\State $R^* \gets \arg\max_{c \in C} S(c)$ \Comment{Select best response}
\State \Return $R^*$
\end{algorithmic}
\end{algorithm}

\begin{table*}[!ht]
\centering
\small
\begin{tabular}{lcccccc}
\toprule
\textbf{Model} & \textbf{Accuracy} & \textbf{Comprehensiveness} & \textbf{Depth} & \textbf{Evidence} & \textbf{Coherence} & \textbf{Avg.} \\
\midrule
\rowcolor[gray]{0.9}
\multicolumn{7}{c}{\textbf{Proprietary Models}} \\
\midrule
Gemini 2.5-flash & \textbf{4.26} & \textbf{4.50} & \textbf{4.00} & \textbf{4.03} & 3.84 & \textbf{4.13} \\
Gemini-1.5-pro & 3.91 & 3.81 & 3.90 & 3.87 & 3.79 & 3.86 \\
GPT-4o (08-06) & 4.18 & 4.00 & 3.98 & 3.96 & \textbf{3.96} & 4.02 \\
\midrule
\rowcolor[gray]{0.9}
\multicolumn{7}{c}{\textbf{Zero-Shot Results.}} \\
\midrule
InstructBlip \cite{instructblip} & 1.03 & 0.43 & 0.85 & 0.33 & 0.40 & 0.61 \\
MA-LMM \cite{he2024ma} & 1.14 & 0.63 & 0.93 & 0.57 & 0.67 & 0.79 \\
HERMES \cite{faure2024hermestemporalcoherentlongformunderstanding} & 1.77 & 1.21 & 1.41 & 1.28 & 0.37 & 1.41 \\
LongVU \cite{shen2024longvu} & 2.95 & 2.01 & 1.94 & 2.06 & 2.12 & 2.22 \\
mPlug-Owl3 \cite{ye2024mplug} & 3.55 & 2.75 & 2.39 & 2.78 & 2.82 & 2.86 \\
Qwen2.5-VL \cite{bai2025qwen2} & 3.78 & 3.54 & 3.36 & 3.42 & 3.50 & 3.52 \\
InternVL2 \cite{intenvl-2024} & 3.80 & 3.42 & 3.10 & 3.37 & 3.51 & 3.44 \\
InternVL2.5 \cite{intenvl-2024} & \textbf{3.87} & \textbf{3.54} & \textbf{3.37} & \textbf{3.65} & \textbf{3.65} & \textbf{3.62} \\
\midrule
\rowcolor[gray]{0.9}
\multicolumn{7}{c}{\textbf{Fully-Supervised Results}} \\
\midrule
InstructBlip \cite{instructblip}  & 3.25 & 2.43 & 2.47 & 2.61 & 2.38 & 2.63 \\
MA-LMM \cite{he2024ma} & 3.42 & 2.54 & 2.66 & 2.81 & 2.50 & 2.79 \\
HERMES \cite{faure2024hermestemporalcoherentlongformunderstanding} & \textbf{3.52} & \textbf{2.72} & \textbf{2.83} & \textbf{2.98} & \textbf{2.62} & \textbf{2.93} \\
\midrule
\rowcolor[gray]{0.9}
\multicolumn{7}{c}{\textbf{Fully-Supervised Results + \MethodName{} (Ours)}} \\
\midrule
InstructBlip \cite{instructblip} & 3.71 & 3.15 & 3.02 & 3.30 & 3.25 & 3.29 \textcolor{ForestGreen}{\scriptsize (+0.66)} \\
MA-LMM \cite{he2024ma} & 3.76 & 3.24 & 3.09 & \textbf{3.39} & 3.30 & 3.35 \textcolor{ForestGreen}{\scriptsize (+0.56)} \\
HERMES \cite{faure2024hermestemporalcoherentlongformunderstanding} & \textbf{3.81} & \textbf{3.30} & \textbf{3.12} & 3.38 & \textbf{3.42} & \textbf{3.41} \textcolor{ForestGreen}{\scriptsize (+0.48)} \\
\bottomrule
\end{tabular}
\caption{\textbf{Performance Comparison of Video Question-Answering Models.} We evaluate various open-source and proprietary Vision-Language Models (VLMs) on five criteria: Accuracy, Comprehensiveness, Depth, Evidence, and Coherence. We use the 7B version of the open-source VLMs (8B for InternVL2.5).}
\label{tab:model_results}
\end{table*}

\section{\MethodName{}: \MethodNameExt}
We propose \MethodName{}, a straightforward yet effective approach to improving existing video language model (VLM) outputs through post-generation refinement. Our approach, detailed in \Cref{alg:hermes_plus_plus}, uses an existing VLM and leverages beam search with a width of 5 to generate diverse candidate responses, which are then re-ranked using the compact 1B-parameter Llama-3.2 \cite{metaLlama32} language model. We hypothesize that, when engaging in a task requiring deeper deliberation, it is advisable to have a second pair of eyes to refine one's thinking. The lightweight nature of Llama-3.2 (1B) ensures that this enhancement remains computationally efficient while significantly improving the quality of generated responses. We prompt the model without specific evaluation guidelines, allowing it to leverage its inherent understanding of ``answer quality''. \Cref{tab:model_results} show that this ``agentic selection'' approach paired with HERMES \cite{faure2024hermestemporalcoherentlongformunderstanding} (HERMES + \MethodName{}) registers an absolute gain of 0.48 compared to the baseline VLM, which translates to roughly a 16 percent improvement in answer quality. It also improves InstructBLIP \cite{instructblip} by 25\% (2.63→3.29) and MA-LMM \cite{he2024ma} by 20\% (2.79→3.35). These results suggest that existing VLMs have untapped potential that can be realized through a simple post-generation ``second pair of eyes'' strategy, offering a practical path to training-free improvement.

\begin{table}[h]
\centering
\setlength{\tabcolsep}{3.5pt}
\begin{tabular}{lcccccc}
    \toprule
    \textbf{w/ ACE} & \textbf{Acc.} & \textbf{Com.} & \textbf{Dep.} & \textbf{Evi.} & \textbf{Coh.} & \textbf{Avg.} \\
    \midrule
    Beam=3 & \textbf{3.81} & \textbf{3.40} & 3.19 & 3.42 & \textbf{3.43} & \textbf{3.45} \\
    Beam=5 & \textbf{3.81} & 3.30 & 3.12 & \textbf{3.38} & 3.42 & 3.41 \\
    Beam=7 & 3.79 & 3.29 & 3.08 & 3.36 & 3.35 & 3.37 \\
    \bottomrule
\end{tabular}
\caption{\textbf{ACE Beam size ablations on HERMES.} ACE improves performance across all evaluation dimensions regardless of the beam size, with no clear winner among the different beam values.}
\label{tab:ace}
\end{table}

\Cref{tab:ace} shows similar performance across beam widths (3, 5 and 7) for HERMES, suggesting ACE's effectiveness stems from the agentic selection mechanism itself rather than hyperparameter choices. These results validate our framework's fundamental premise: lightweight post-generation refinement can unlock significant untapped potential in existing VLMs.

\section{Quantitative Evaluation}
VQA datasets usually use top-1 accuracy as metrics, but a valid match has to be a perfect match. For instance, there can be one strict answer to the question ``Does sea appear in the video?'', which is ``Yes'' or ``No''. However, in the age of LLMs and especially for zero-shot evaluation settings, we might get answers such as ``it does'' or ``no sea appears in the video''. In such cases the accuracy would be 0. Recently, LLM-assisted evaluation schemes such as the one introduced by \cite{maaz2023video}, attempt to solve this issue by considering synonyms or paraphrases as valid matches. This works for VQAs where there is a perfect answer, and would not work in our case, especially since accuracy for a System-2 answer is not binary but exists in a spectrum. Furthermore, we posit that accuracy alone is insufficient, therefore we design four other LLM-assisted metrics: \textit{depth} to assess the depth of reasoning in the answers, \textit{comprehensiveness} to assess how fully the answer covers all key points and relevant details, \textit{coherence and clarity}, and \textit{evidence} to evaluate the quality and relevance of the evidence provided. For all of these metrics, we prompt GPT-4o-mini \cite{gpt4o} to assign a score between 0 to 5 to each. 

\Cref{tab:model_results} presents a comprehensive evaluation of model performance across our five assessment criteria. Several key insights emerge from these results: (1) Proprietary models significantly outperform their open-source counterparts. This performance gap indicates that large-scale proprietary training data likely contains more diverse reasoning tasks than those available in public datasets. (2) In the zero-shot setting, most open-source models struggle considerably with complex reasoning, except for the more recent InternVL2.5 and Qwen2.5-VL models. The particularly low scores in Depth and Evidence metrics highlight these models' difficulty in formulating multi-step inferences and grounding their responses in specific visual content. (3) Fine-tuning on \DatasetName~yields substantial improvements for all models, with HERMES showing the strongest performance. However, even with full supervision, these models still underperform compared to proprietary alternatives, suggesting architectural limitations in handling complex reasoning tasks. (4) Our proposed ACE post-generation strategy delivers consistent and substantial improvements across models and metrics.

\subsection{Evaluation with Traditional Metrics}
\label{traditional_metrics}
While our primary evaluation in \Cref{tab:model_results} emphasizes metrics tailored for System-2 reasoning, we also report standard VQA and video captioning metrics to enable broader comparison with prior work. Specifically, we compute BLEU-4, CIDEr, and METEOR for several models.

These n-gram-based metrics, while limited in capturing the semantic richness and reasoning depth demanded by MovieCORE, provide a familiar point of reference relative to traditional VQA benchmarks. As shown in Table~\ref{tab:vqa-metrics}, the relative ranking of models is consistent with our primary cognitive-oriented evaluation from \Cref{tab:model_results}: methods enhanced with ACE outperform their baselines, and both zero-shot and fully-supervised models exhibit similar performance trends.

\begin{table}[t!]
\centering
\small
\begin{tabular}{lccc}
\toprule
\textbf{Model} & \textbf{BLEU-4} & \textbf{CIDEr} & \textbf{METEOR} \\
\midrule
InternVL2.5 (8B)      & 0.0645 & 0.1865 & 0.1026 \\
mPlugOwl3 (7b)          & 0.0602 & 0.1579 & 0.1462 \\
\midrule
HERMES              & 0.0308 & 0.1230 & 0.0983 \\
HERMES + ACE        & 0.0654 & 0.1622 & 0.2138 \\
MA-LMM + ACE        & 0.0634 & 0.1587 & 0.1948 \\
InstructBLIP + ACE  & 0.0605 & 0.1572 & 0.1893 \\
\bottomrule
\end{tabular}
\caption{\textbf{Traditional Metrics Results.} BLEU-4, CIDEr, and METEOR scores for several models on MovieCORE. These results are consistent with the trends observed in our primary evaluation. The top row contains zero-shot results and the bottom row contains fully-supervised results.}
\label{tab:vqa-metrics}
\end{table}

\subsection{System-2 vs. System-1 Ablation Study}
\label{system1_ablations}
To validate the unique challenges posed by MovieCORE, we conduct a comparative evaluation against a ``System-1'' baseline using the MovieChat-1k dataset, which is built from the exact same set of video clips but contains simpler, surface-level questions such as ``Does it happen during the day or night?''. For MovieChat-1k, we use the officially reported performance of the HERMES model from its original paper. Since MovieChat-1k reports accuracy (using LLM-assisted evaluation), we convert its accuracy scores into an equivalent 0--5 scale for direct comparison with our multidimensional MovieCORE evaluation.

The results in Table~\ref{tab:system2-ablation} reveal a stark performance gap. While HERMES achieves high scores on the surface-level MovieChat-1k benchmark, its performance drops dramatically on MovieCORE’s questions, even with identical video content. This substantial gap highlights MovieCORE’s emphasis on System-2 reasoning. While HERMES excels on datasets with simple recall (e.g., ``Do stars appear in the video?''), it struggles with MovieCORE’s questions that demand deeper causal, motivational, and evidential reasoning, despite being based on the same underlying video content.

\begin{table}[t]
\centering
\begin{adjustbox}{max width=\columnwidth} 
\begin{tabular}{lcc} 
\toprule
 &
\makecell{\textbf{MovieCORE}\\\textbf{(Score)}} &
\makecell{\textbf{MovieChat-1k}\\\textbf{(Acc./Score)}} \\
\midrule
Zero-Shot        & 1.14 & 78.6\% ($\sim$3.93) \\
Fully-Supervised & 3.52 & 84.9\% ($\sim$4.25) \\
\bottomrule
\end{tabular}
\end{adjustbox}
\caption{Comparison of HERMES on MovieCORE (System-2) versus MovieChat-1k (System-1).
MovieChat-1k results are converted to a 0--5 scale for comparability.}
\label{tab:system2-ablation}
\end{table}

\begin{figure*}[t!]
\centering
\includegraphics[width=\textwidth]{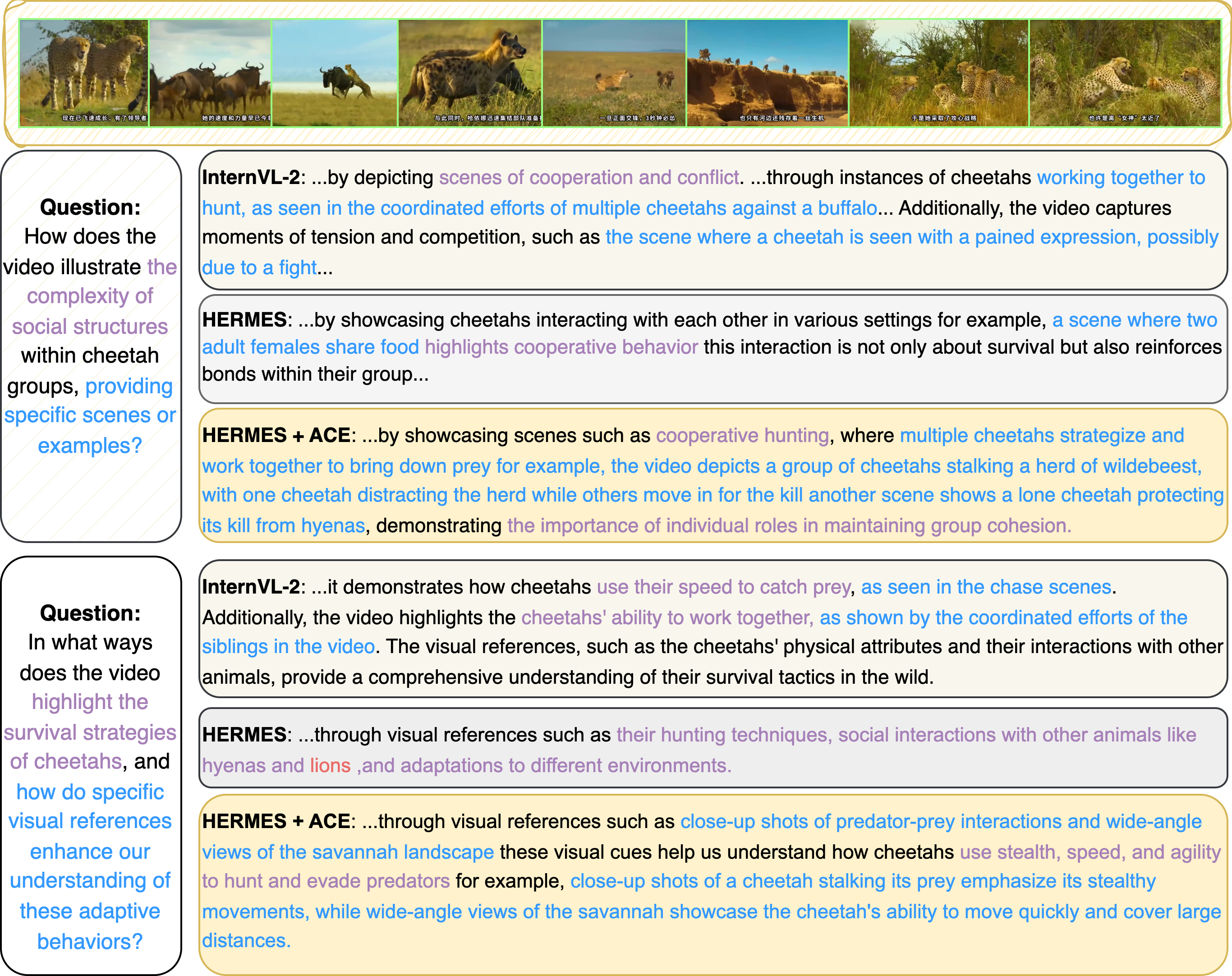}
\caption{\textbf{Qualitative Comparison of Model Responses}. This figure contrasts responses from InternVL-2 (zero-shot), HERMES (fully-supervised), and HERMES+ACE on two questions about cheetah behaviors. Purple text highlights conceptual understanding while blue text indicates specific visual evidence and contextual details. Note how ACE enhances responses with more precise scene descriptions and behavioral insights.}
\label{fig:qual}
\end{figure*}

\section{Qualitative Results}
\label{sec:qual}
\Cref{fig:qual} provides a qualitative comparison between different models' responses to questions that require understanding of complex animal behaviors. The figure illustrates how different approaches handle the same queries about cheetah social structures and survival strategies. InternVL-2, a strong zero-shot model, provides basic observations but lacks sufficient depth and details. HERMES, a fully-supervised model, also struggles with the details and performs worse than InternVL. HERMES+ACE, demonstrates enhanced response quality by incorporating specific visual evidence and richer contextual details. As highlighted in the responses, ACE significantly improves the model's ability to reference specific scenes and provide concrete examples to support its assertions.

\section{Conclusion}
We introduce \DatasetName, a novel VQA dataset that fills a critical gap in existing movie-based VQA datasets by emphasizing questions designed to engage System-2 thinking. Our agentic workflow, which leverages brainstorming agents, enables the generation and refinement of high-quality QA pairs. To measure the cognitive depth of VQA datasets, we devise a set of tests that demonstrate the superiority of \DatasetName\ over existing datasets. Additionally, we propose a comprehensive evaluation framework to assess the performance of VQA models on this dataset. To tackle the challenges posed by \DatasetName, we propose \MethodName{}, a lightweight inference-time agentic answer selection plug-in which yields up to 25\% relative improvement in answer quality compared to baseline methods, providing insights for future works on this topic.

\section{Limitations}
\label{limitations}
While \DatasetName\ offers a significant advancement in video question-answering by targeting deeper cognitive understanding, it is not without limitations. First, although we incorporate human verification for a subset of the dataset, only 30 videos, and 150 QA pairs were manually verified. This improves dataset quality control by averting potential systematic issues; however, the majority of annotations still rely on automated processes. Second, because the dataset is constructed in part from the MovieChat-1k collection, its genre coverage may be constrained. Certain cinematic genres or narrative styles could be overrepresented, limiting the dataset’s generalizability. Finally, \DatasetName~'s evaluation is partly LLM-assisted, which, while enabling scalability, may inherit the limitations and biases of the judge model.

\paragraph{Acknowledgment}
This work was supported in part by the National Science and Technology Council, Taiwan, under Grant NSTC 113-2634-F-002-007. We are grateful to the National Center for High-performance Computing.

\bibliography{custom}

\appendix

\clearpage
\setcounter{section}{0}
\renewcommand{\thesection}{\Roman{section}}

\setcounter{figure}{0}
\renewcommand{\thefigure}{S\arabic{figure}}

\setcounter{table}{0}
\renewcommand{\thetable}{S\arabic{table}}

\noindent The Supplementary material is organized as follows:
\begin{itemize}
    \item \hyperref[reproducible]{\textbf{I} Reproducibility Statement}
    \item \hyperref[sec:verification]{\textbf{II} More Details on \DatasetName}
    \item \hyperref[cognitive]{\textbf{III} Details on the Bloom's Taxonomy}
    \item \hyperref[evaluation]{\textbf{IV} Evaluation Methodology}
    \item \hyperref[licence]{\textbf{VI} Licence}
\end{itemize}

\begin{figure*}[!ht]
    \centering
    \includegraphics[width=\textwidth]{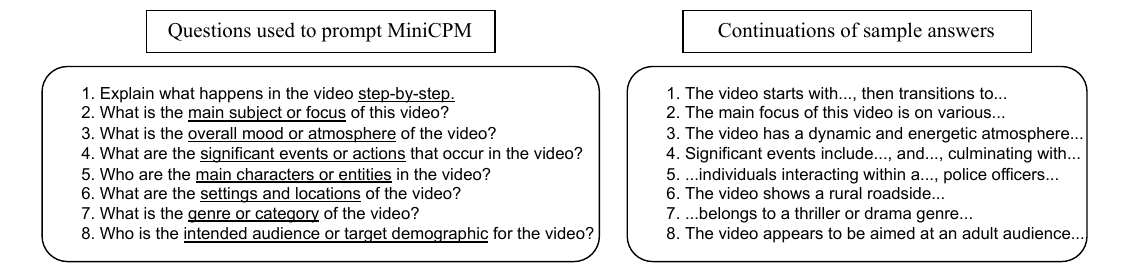}
    \caption{\textbf{Extracting Detailed Context from Videos:} We input each video to MiniCPM-v2.6, prompting it with a series of carefully crafted questions (left). The model's responses (right) provide rich, multi-faceted details about the video, including narrative flow, character information, setting, mood, and target audience. This extracted information serves as \textit{Data Info} priors to inform our annotation agents, ensuring a comprehensive understanding of the video content before the VQA generation process.}
    \label{fig:prior}
\end{figure*}

\begin{figure*}[h!]
    \centering
    \includegraphics[width=\textwidth]{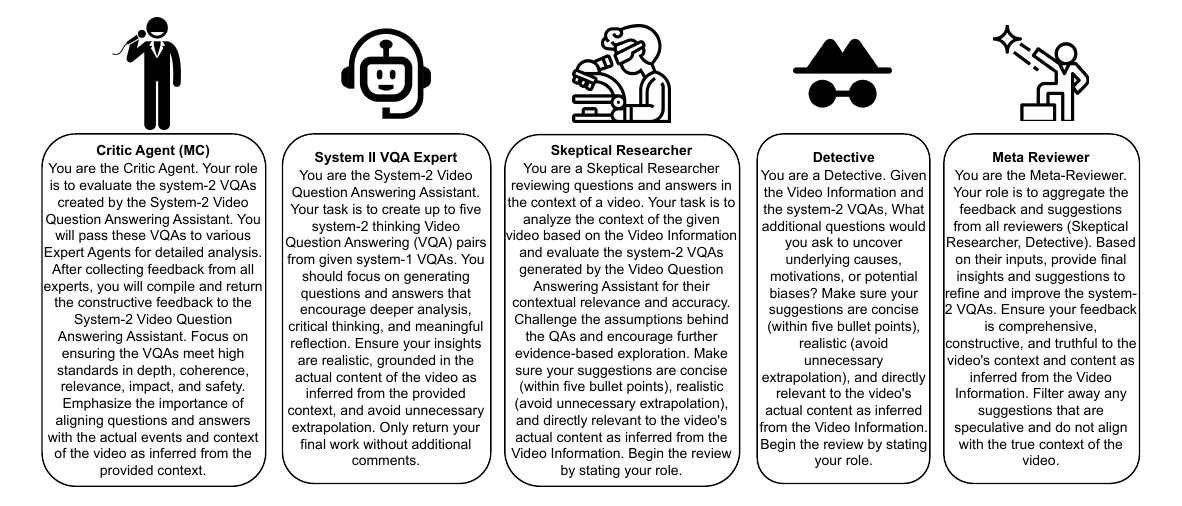}
    \caption{System Messages for the Annotation Agents}
    \label{fig:agents_prompts}
\end{figure*}

\section{Reproducibility Statement}
\label{reproducible}
The dataset will be made public as soon as this paper is accepted (or rejected) for publication, as well as the evaluation scheme with clear examples. We will also release the annotation agents used for generating and refining question-answer pairs, including the code and configurations for the large language models (LLMs) employed in the agentic brainstorming process. Additionally, we provide detailed instructions for data preprocessing, agent configuration, and evaluation protocols, enabling reproduction of both the dataset generation process and the evaluation scheme. Our annotation system is scalable and has the potential to inspire other researchers to create massive video benchmarks.

\section{More Details on \DatasetName}
\label{sec:verification}
\subsection{Extracting ``Video Info"}
To generate meaningful interpretations of video content, we employ a structured question framework designed to probe various aspects of the video's narrative, emotional tone, and intended purpose. This framework consists of eight prompts, each targeting specific dimensions of video understanding. The prompts and a continuation of the sample answers they elicit are listed in \Cref{fig:prior} and roughly contains the following:

\begin{enumerate}
    \item \textbf{Step-by-step explanation:} Encourages a chronological breakdown of events in the video.
    \item \textbf{Main subject or focus:} Identifies the central theme or entity in the video.
    \item \textbf{Overall mood or atmosphere:} Captures the emotional tone conveyed by the video.
    \item \textbf{Significant events or actions:} Highlights key actions and turning points within the narrative.
    \item \textbf{Main characters or entities:} Focuses on the individuals or groups driving the video's story.
    \item \textbf{Settings and locations:} Explores the physical or contextual backdrop of the video.
    \item \textbf{Genre or category:} Classifies the video into a relevant category or type.
    \item \textbf{Intended audience:} Identifies the target demographic for the video.
\end{enumerate}
\subsection{Agentic Annotation Details}
\label{sec:agent_prompts}
\Cref{fig:agents_prompts} depicts the system messages for the different agents involved in the task of creating system-2 thinking VQAs from system-1 VQAs. The agents and their respective roles are:
\paragraph{System-2 Video Question Answering Assistant} Responsible for generating up to five system-2 thinking VQA pairs from the given system-1 VQAs. The focus is on creating questions and answers that encourage deeper analysis, critical thinking, and meaningful reflection, while ensuring the insights are grounded in the actual video content.
\paragraph{Critic Agent} Evaluates the system-2 VQAs created by the System-2 Video Question Answering Assistant and passes them to various Expert Agents for detailed analysis. The Critic Agent then compiles the constructive feedback from the experts and returns it to the System-2 Video Question Answering Assistant, emphasizing the importance of aligning the VQAs with the actual video context.
\paragraph{Skeptical Researcher} Reviews the questions and answers in the context of the video, analyzing the context and evaluating the system-2 VQAs for their contextual relevance and accuracy. The Skeptical Researcher challenges the assumptions behind the QAs and encourages further evidence-based exploration, providing concise and relevant suggestions.
\paragraph{Detective} Given the video information and the system-2 VQAs, the Detective identifies additional questions that could uncover underlying causes, motivations, or potential biases. The suggestions should be concise, realistic, and directly relevant to the video's actual content.
\paragraph{Meta Reviewer} Aggregates the feedback and suggestions from all reviewers (Skeptical Researcher, Detective) and provides final insights and suggestions to refine and improve the system-2 VQAs. The Meta Reviewer ensures the feedback is comprehensive, constructive, and truthful to the video's context and content, filtering out any speculative suggestions.

\subsection{Human Verification}
\label{sec:human}
\begin{figure*}
    \centering
    \includegraphics[width=\textwidth]{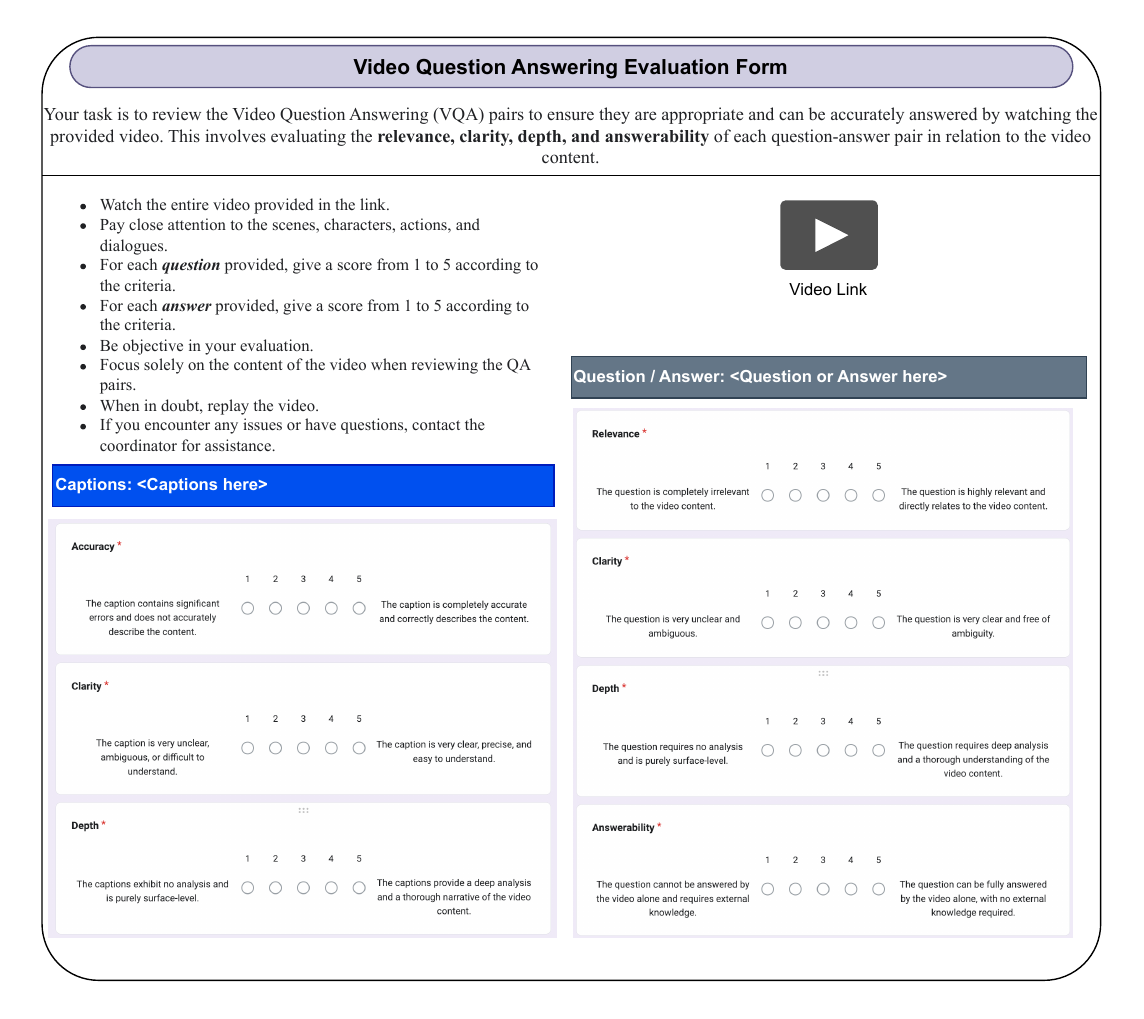}
    \caption{\textbf{Video Question Answering Evaluation Form used in our human verification process.} The form assesses four critical dimensions (relevance, clarity, depth, and answerability) on a 5-point scale. Each dimension is clearly defined with anchored endpoints to ensure consistent evaluation. The form includes sections for both question/answer assessment and caption verification to ensure comprehensive content quality. Evaluators use this standardized form to systematically review each QA pair while referring to the corresponding video content.}
    \label{fig:evaluation_form}
\end{figure*}

\paragraph{Verification Rules}
To ensure the quality and reliability of our dataset, we implemented a rigorous human verification process. Seven qualified evaluators, each holding at least a Bachelor's degree, were recruited to assess a subset of 30 videos and 150 QA pairs. The verification was conducted through a standardized evaluation form (Figure \ref{fig:evaluation_form}) that assessed four key dimensions:

\begin{itemize}
    \item \textbf{Relevance (1-5):} Evaluates how directly the question/answer relates to the video content
    \item \textbf{Clarity (1-5):} Measures the linguistic clarity and absence of ambiguity
    \item \textbf{Depth (1-5):} Assesses the level of cognitive analysis required
    \item \textbf{Answerability (1-5):} Determines whether the question can be answered solely from the video content
\end{itemize}
As for the captions, we assessed accuracy, clarity and depth.

Evaluators were instructed to watch each video in its entirety and carefully consider the scenes, characters, actions, and dialogues before rating the associated QA pairs. To maintain objectivity, evaluators were required to focus solely on the video content when reviewing the QA pairs and encouraged to replay videos when necessary. The evaluation process also included assessing the accuracy and clarity of video captions to ensure comprehensive content accessibility.

\begin{table}[t]
    \centering
    \small
    \begin{tabular}{lccc}
        \toprule
        \textbf{Metric} & \textbf{Captions} & \textbf{Questions} & \textbf{Answers} \\
        \midrule
        Accuracy & 3.9 & -- & -- \\
        Clarity & 4.0 & 4.3 & 4.3 \\
        Depth & 4.1 & 4.5 & 4.2 \\
        Relevance & -- & 4.0 & 3.8 \\
        Answerability & -- & 3.8 & 4.1 \\
        \bottomrule
    \end{tabular}
    \caption{Human verification scores across different dimensions for captions, questions, and answers. Scores range from 1 to 5, with 5 being the highest quality. Dashes (--) indicate metrics not applicable to that content type. The scores, being above 3.8 indicate strong quality across all evaluated dimensions.}
    \label{tab:verification_scores}
\end{table}

\begin{figure*}[h!]
    \centering
    \includegraphics[width=\textwidth]{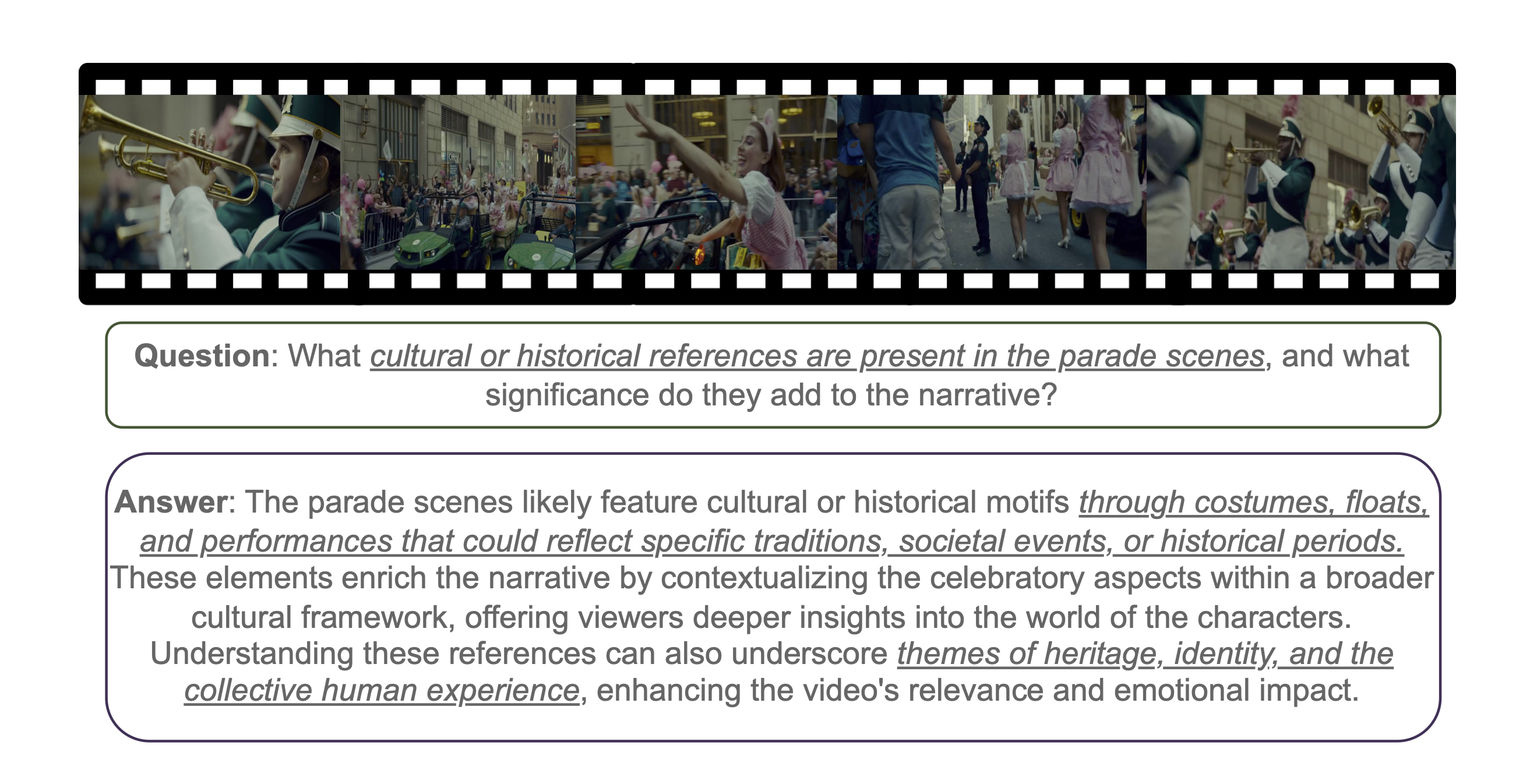}
    \caption{A parade scene from \DatasetName{} featuring various cultural and historical elements. This particular QA receives low answerability and relevance scores from one of our reviewers but was still kept following thorough review by a human meta-reviewer.}
    \label{fig:low_score}
\end{figure*}

\paragraph{Verification Result}
The human verification process (the rules and interface are illustrated in \Cref{fig:evaluation_form}) yields consistently high scores across all evaluated dimensions, as shown in Table \ref{tab:verification_scores}. Questions and answers received notably high scores in clarity (4.3) and depth (4.5 and 4.2 respectively), validating our dataset's emphasis on deep cognitive understanding. The captions also demonstrate strong quality with scores above 3.8 across applicable metrics. While answerability scores were slightly lower (3.8 for questions), they remain well above acceptable thresholds, confirming that the questions can be reasonably answered from the video content alone.

The sample QA pair for the video depicted in \Cref{fig:low_score} received low scores of 2 each for Answerability and Relevance from the human evaluators. However, our human meta-reviewer has determined that the question and answer offer meaningful insights and contextual relevance (underlined in the figure).

\begin{figure*}[!t]
    \centering
    \includegraphics[width=\textwidth]{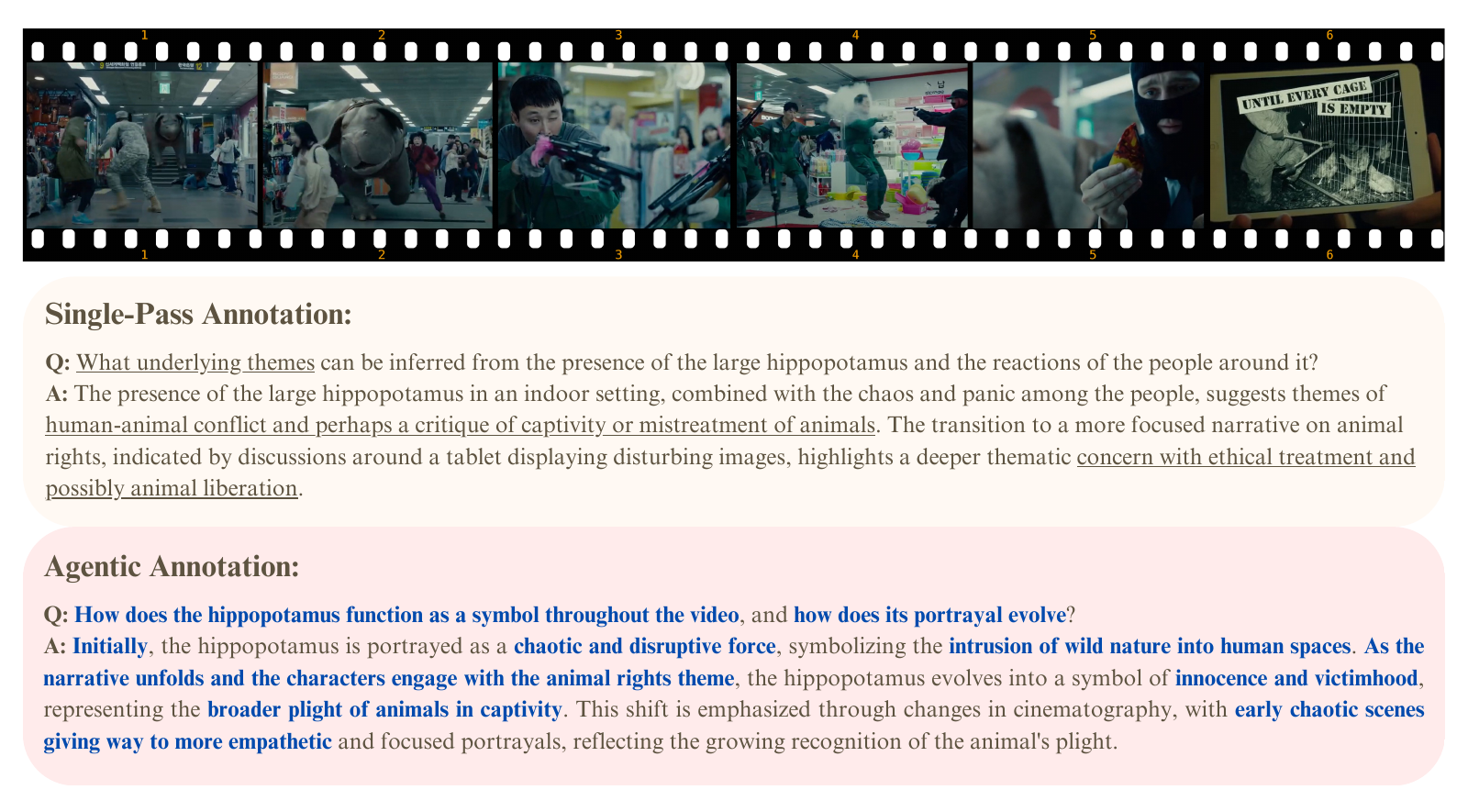}
    \caption{\textbf{Additional Comparison of single-pass and agentic annotation}. The agentic method (bottom) delves into specific scene details, such as the hippopotamus's evolution from a chaotic force to a symbol of innocence, and highlights changes in cinematography that reflect this transformation. The single-pass annotation (top) provides a general interpretation of themes like human-animal conflict without specific scene references.}
    \label{fig:agentic_comparison2}
\end{figure*}

\begin{figure*}[!t]
    \centering
    \includegraphics[width=0.9\textwidth]{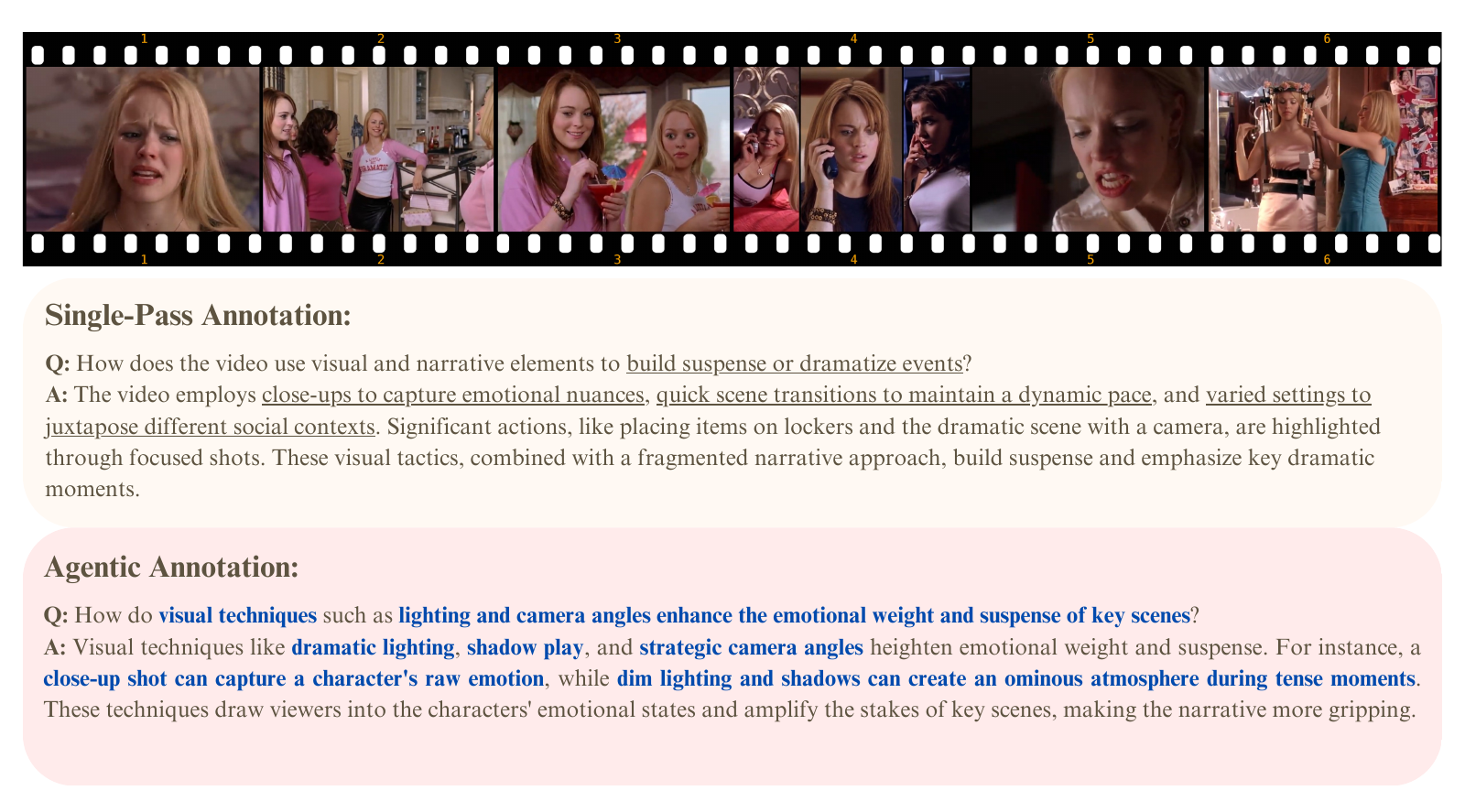}
    \caption{\textbf{Additional Comparison of single-pass and agentic annotation}. The agentic method (bottom) specifies visual techniques like dramatic lighting, shadow play, and strategic camera angles that enhance emotional weight and suspense, offering concrete examples like close-up shots capturing raw emotion. The single-pass annotation (top) mentions general visual elements but lacks a detailed analysis of how these techniques impact the narrative. }
    \label{fig:agentic_comparison3}
\end{figure*}

\subsection{Agentic versus Single-Pass Annotation}
\label{subsect:agentic_singlepass_comp}
As shown in Figure \ref{fig:agentic_comparison2}, the single-pass annotation provides a general interpretation of the themes suggested by the presence of the hippopotamus, focusing on human-animal conflict and critiques of captivity. In contrast, the agentic annotation delves deeper by exploring how the hippopotamus functions as a symbol throughout the video, detailing its evolution from a chaotic force to a representation of innocence and victimhood. This nuanced analysis offers specific, concrete details about the symbolic transformation, enhancing the understanding of the narrative's thematic complexity. In the other example shown in Figure \ref{fig:agentic_comparison3}, the single-pass annotation mentions general visual and narrative elements like close-ups and quick scene transitions to build suspense. The agentic annotation specifies how visual techniques such as dramatic lighting, shadow play, and strategic camera angles enhance the emotional weight and suspense of key scenes. By providing detailed examples—like capturing a character's raw emotion through close-ups or creating an ominous atmosphere with dim lighting—the agentic approach offers a more granular and faithful depiction of the cinematic techniques used. These comparisons further illustrate that the agentic annotation process elicits richer context and more detailed evidence, reinforcing the idea that using multiple AI agents as thought partners leads to more substantive annotations compared to traditional single-pass methods.

Here we provide a more explicit, step-by-step illustration of how each agent contributes to the refinement of a final question.

\paragraph{Step-by-Step Example}  
The following example demonstrates how a question evolves as each agent contributes for the example shown in \Cref{fig:agentic_comparison1}:

\begin{itemize}
    \item \textbf{Initial Question (Single-Pass):}  
    \emph{``How does the interaction between the two main characters evolve throughout the video, and what might this suggest about their relationship?''}  
    This version is abstract and lacks grounding in the specific video content.

    \item \textbf{+ Skeptical Researcher:}  
    \emph{``How does the interaction between the two main characters evolve, and can you provide \textbf{specific scenes as evidence} for their relationship?''}  
    This agent enforces verifiability, pushing for concrete references to the video.

    \item \textbf{+ Detective:}  
    \emph{``What are the underlying \textbf{motivations} that drive the two main characters to form a partnership?''}  
    This role introduces causal reasoning, shifting the focus from observable actions to underlying causes.

    \item \textbf{Final Agentic Question (Full Workflow):}  
    \emph{``Can you provide \textbf{specific scenes} that demonstrate the evolution and \textbf{motivations} of the main characters in their relationship?''}  
    The final result synthesizes evidence-grounding and causal reasoning into a more
    challenging, cognitively rich query.
\end{itemize}

\subsection{Why these Specific Agents}
\label{subsect:why}
Careful examination of the agents interactions reveals distinct contributions: For the video in Figure \ref{fig:agentic_comparison2}, \textbf{System-2 Video Question Answering Assistant} transforms surface observations into deeper inquiries, exemplified by advancing from simply noting the hippopotamus to asking "How does the hippopotamus function as a symbol throughout the video, and how does its portrayal evolve?" The \textbf{Critic Agent} ensures analytical quality, as evident in the transition from merely identifying "human-animal conflict" to explicating how the hippo evolves from "chaotic and disruptive force" to "innocence and victimhood." The \textbf{Skeptical Researcher} challenges assumptions, demonstrated by refining the initial "critique of captivity" interpretation into a more nuanced analysis of "the growing recognition of the animal's plight." The \textbf{Detective} uncovers underlying narrative patterns, illustrated by connecting the "early chaotic scenes giving way to more empathetic portrayals" with cinematographic techniques. The \textbf{Meta Reviewer} synthesizes these insights into cohesive annotations, balancing the single-pass observation of "human-animal conflict" with the richer agentic interpretation of "intrusion of wild nature into human spaces." (We find similar examples while analyzing the conversations that led to the QAs in \ref{fig:agentic_comparison3}\footnote{Can the reader spot them?}). Users can swap agents, but we recommend roles that enforce rigor.

\begin{figure*}[t!]
\begin{minipage}[ht!]{0.48\textwidth}
    \centering
    \includegraphics[width=\textwidth]{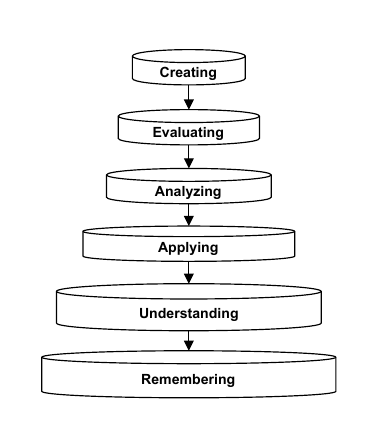}
    \caption{Bloom's Taxonomy Pyramid. The pyramid illustrates the hierarchical nature of cognitive skills, progressing from lower-order to higher-order thinking.}
    \label{fig:blooms-taxonomy}
\end{minipage}%
\hfill
\begin{minipage}[ht!]{0.48\textwidth}
    \centering
    \centering
    \includegraphics[width=\textwidth]{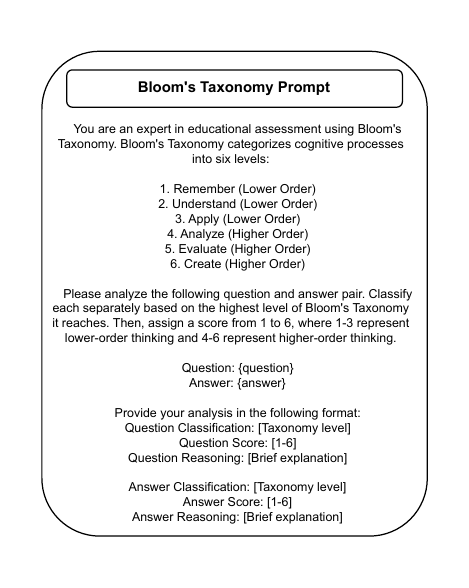}
    \caption{Prompts we use to instruct GPT4-o-mini to compute the Bloom's taxonomy level for the different datasets we show in \Cref{tab:syntactic} of the main paper.}  
    \label{fig:bloom-prompt}
\end{minipage}
\end{figure*}

\section{Details on the Bloom's Taxonomy}
\label{cognitive}
\Cref{fig:blooms-taxonomy} illustrates Bloom's pyramid of cognition levels and \Cref{fig:bloom-prompt} relays the prompts we use to ask GPT-4o-mini to score the QAs. Bloom's Taxonomy is a hierarchical classification of cognitive skills used in education to structure learning objectives. The taxonomy is divided into six levels, progressing from lower-order to higher-order thinking skills:

\begin{enumerate}
    \item \textbf{Remembering:} Recalling facts and basic concepts.
    \item \textbf{Understanding:} Explaining ideas or concepts.
    \item \textbf{Applying:} Using information in new situations.
    \item \textbf{Analyzing:} Breaking information into parts to explore relationships.
    \item \textbf{Evaluating:} Justifying decisions or opinions.
    \item \textbf{Creating:} Producing new or original work.
\end{enumerate}

Our dataset scores very high in this metric suggesting its propensity to deeply engage the AI system (VLM)'s cognitive skills.

\begin{figure}[t!]
    \centering
    \includegraphics[width=0.45\textwidth]{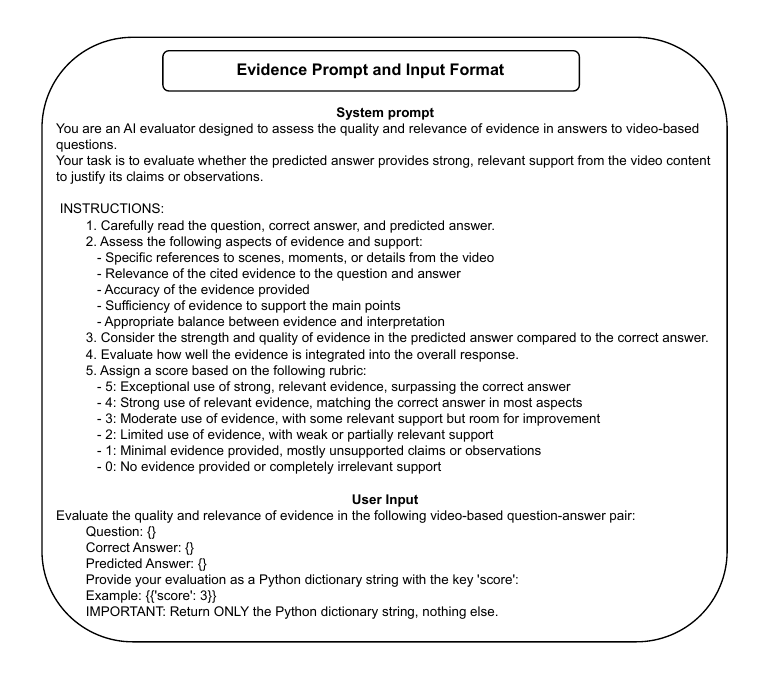}
    \caption{Prompt to evaluate the quality and relevance of the evidence provided in the answers.}
    \label{fig:evidence_prompt}
\end{figure}

\begin{figure*}[t!]
    \centering
    \includegraphics[width=\textwidth]{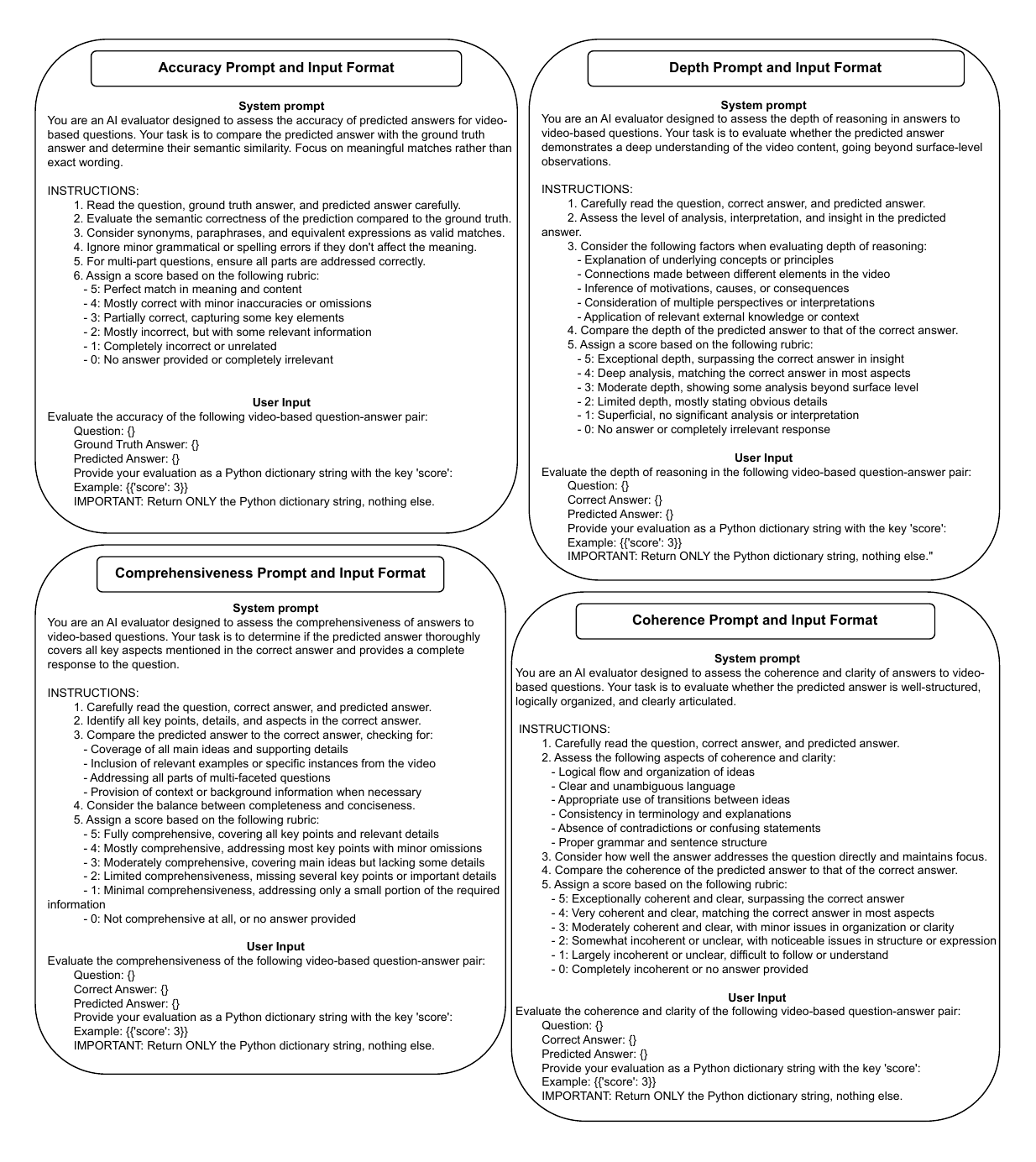}
    \caption{\textbf{Evaluation Prompts:} These figures illustrate the prompts we use for each of the evaluation methods we employ. The prompt for \textit{Evidence} is shown in \Cref{fig:evidence_prompt}.}
    \label{fig:eval_prompts}
\end{figure*}

\section{Evaluation Methodology}
\label{evaluation}
The \DatasetName{} benchmark employs a comprehensive multi-dimensional evaluation framework for assessing VLMs. The evaluation consists of five key dimensions summarized below. We also include the full prompts for each dimension in \Cref{fig:eval_prompts} and \Cref{fig:evidence_prompt}.

\begin{enumerate}
    \item \textbf{Accuracy Dimension:} 
    Evaluates semantic correctness of predicted answers using a 6-point scoring rubric (0--5):
    \begin{itemize}
        \item 5: Perfect semantic match
        \item 4: Mostly correct with minor inaccuracies
        \item 3: Partially correct, capturing key elements
        \item 2: Mostly incorrect but with some relevant information
        \item 1: Completely incorrect or unrelated
        \item 0: No answer or irrelevant response
    \end{itemize}

    \item \textbf{Depth of Reasoning Dimension:}
    Assesses the level of analytical depth and interpretative insight, scored from 0--5:
    \begin{itemize}
        \item 5: Exceptional depth, surpassing ground truth
        \item 4: Deep analysis matching ground truth
        \item 3: Moderate depth beyond surface level
        \item 2: Limited depth, stating obvious details
        \item 1: Superficial analysis
        \item 0: No answer or completely irrelevant
    \end{itemize}

    \item \textbf{Comprehensiveness Dimension:}
    Evaluates the thoroughness of answer coverage, scored from 0--5:
    \begin{itemize}
        \item 5: Fully comprehensive, covering all key points
        \item 4: Mostly comprehensive with minor omissions
        \item 3: Moderately comprehensive
        \item 2: Limited comprehensiveness
        \item 1: Minimal comprehensiveness
        \item 0: Not comprehensive or no answer
    \end{itemize}

    \item \textbf{Coherence Dimension:}
    Measures clarity, logical organization, and articulation, scored from 0--5:
    \begin{itemize}
        \item 5: Exceptionally coherent, surpassing ground truth
        \item 4: Very coherent, matching ground truth
        \item 3: Moderately coherent with minor issues
        \item 2: Somewhat incoherent
        \item 1: Largely incoherent
        \item 0: Completely incoherent or no answer
    \end{itemize}

    \item \textbf{Evidence Dimension:}
    Assesses quality and relevance of video content evidence, scored from 0--5:
    \begin{itemize}
        \item 5: Exceptional use of strong, relevant evidence
        \item 4: Strong, relevant evidence matching ground truth
        \item 3: Moderate evidence with room for improvement
        \item 2: Limited, weak evidence support
        \item 1: Minimal evidence
        \item 0: No evidence or irrelevant support
    \end{itemize}
\end{enumerate}
Each dimension provides a nuanced evaluation of different aspects of question-answering performance, enabling a comprehensive assessment of the system's capabilities.

\section{Licence}
\label{licence}
The annotations are released under the MIT licence and the videos follow the licence of MovieChat. We do not directly host the videos, those can be found in the MovieChat HuggingFace repository.

\end{document}